\documentclass[11pt]{article}

\usepackage{acl}
\usepackage{times}
\usepackage{latexsym}

\usepackage[T1]{fontenc}
\usepackage[utf8]{inputenc}

\usepackage{microtype}

\usepackage{inconsolata}

\usepackage{graphicx}
\usepackage{amsmath}
\usepackage{tabularx}
\usepackage{multirow}
\usepackage{booktabs}
\usepackage{arydshln}
\usepackage{amssymb}  % 用于三角形符号 \blacktriangle, \blacktriangledown
\usepackage{xcolor}
\definecolor{softgreen}{RGB}{119, 221, 119}
\definecolor{softred}{RGB}{255, 105, 97}
\newcommand{\up}[1]{\textcolor{softgreen}{\blacktriangle\,#1}}
\newcommand{\down}[1]{\textcolor{softred}{\blacktriangledown\,#1}}
\DeclareSymbolFont{extraup}{U}{zavm}{m}{n}
\DeclareMathSymbol{\varheart}{\mathalpha}{extraup}{86}

\title{ConMax: Confidence-Maximizing Compression for Efficient Chain-of-Thought Reasoning}
\author{
  Minda Hu$^{\clubsuit\spadesuit}$\thanks{\ \ The first three authors have equal contributions.}, Zexuan Qiu$^{\clubsuit\spadesuit}$\footnotemark[1], Zenan Xu$^{\spadesuit}$\footnotemark[1], Kun Li$^{\clubsuit\spadesuit}$\\
  \bf
  Bo Zhou$^\spadesuit$$\thanks{B. Zhou and I. King are the corresponding authors.}$, Irwin King$^\clubsuit$$^{\dag}$\\
  $^\clubsuit$The Chinese University of Hong Kong, $^\spadesuit$LLM Department, Tencent\hspace{0.3cm} \\
  \texttt{chaysezhou@tencent.com, king@cse.cuhk.edu.hk}
  }

\begin{document}
\maketitle
\thispagestyle{firstpage}

\begin{abstract}
%\vspace{10cm}
Recent breakthroughs in Large Reasoning Models (LRMs) have demonstrated that extensive Chain-of-Thought (CoT) generation is critical for enabling intricate cognitive behaviors, such as self-verification and backtracking, to solve complex tasks. However, this capability often leads to ``overthinking'',  where models generate redundant reasoning paths that inflate computational costs without improving accuracy. While Supervised Fine-Tuning (SFT) on reasoning traces is a standard paradigm for the 'cold start' phase, applying existing compression techniques to these traces often compromises logical coherence or incurs prohibitive sampling costs. In this paper, we introduce ConMax (Confidence-Maximizing Compression), a novel reinforcement learning framework designed to automatically compress reasoning traces while preserving essential reasoning patterns. ConMax formulates compression as a reward-driven optimization problem, training a policy to prune redundancy by maximizing a weighted combination of answer confidence for predictive fidelity and thinking confidence for reasoning validity through a frozen auxiliary LRM. Extensive experiments across five reasoning datasets demonstrate that ConMax achieves a superior efficiency-performance trade-off. Specifically, it reduces inference length by 43\% over strong baselines at the cost of a mere 0.7\% dip in accuracy, proving its effectiveness in generating high-quality, efficient training data for LRMs.
% \md{Recent breakthroughs in Large Reasoning Models (LRMs) have demonstrated that extensive CoT generation is critical for solving complex tasks. However, this capability often leads to "overthinking," where models generate redundant reasoning paths, inflating computational costs without improving accuracy. While distilling these traces is a standard paradigm for the "cold start" of supervised fine-tuning, existing compression techniques frequently compromise logical coherence or incur prohibitive sampling costs. In this paper, we introduce ConMax (Confidence-Maximizing Compression), a novel reinforcement learning framework designed to automatically compress reasoning traces while preserving essential cognitive behaviors such as self-verification and sub-goal decomposition. ConMax formulates compression as a reward-driven optimization problem, training a policy to prune redundancy by maximizing a weighted combination of Answer Confidence and Thinking Confidence as evaluated by a frozen auxiliary LRM. Extensive experiments across five reasoning datasets demonstrate that ConMax significantly enhances efficiency without sacrificing performance. Our method reduces inference length by 43\% compared to strong baselines with a negligible accuracy drop of 0.7\%, establishing ConMax as a superior strategy for generating high-quality, efficient training data for LRMs.}
\end{abstract}

\section{Introduction}
% Background 
Recent breakthroughs in Large Reasoning Models (LRMs)~\citep{guo2025deepseek,comanici2025gemini,jaech2024openai} have showcased exceptional proficiency in tackling challenging tasks such as math reasoning and code competition. Central to this capability is the utilization of long Chain-of-Thought (CoT), where models generate extensive token sequences to enable intricate reasoning behaviors such as self-verification, backtracking, and error correction~\citep{gandhi2025cognitive,cen2025behavior}. However, this performance gain frequently entails a compromise in efficiency. LRMs often exhibit ``overthinking''~\citep{DBLP:journals/corr/abs-2412-21187}, generating excessively verbose reasoning paths manifested as redundant validation and over-exploration. Crucially, this verbosity often yields diminishing returns for final accuracy~\citep{fan2025missing} while substantially inflating computational overhead and latency.

\begin{figure}[t]
    \centering
    \includegraphics[width=0.96\linewidth]{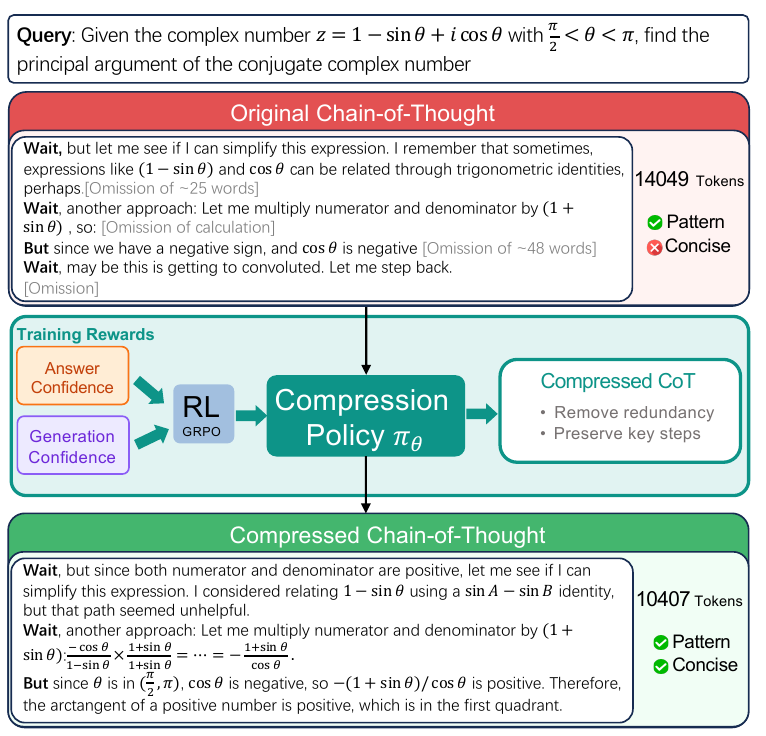}
    \caption{Comparison of reasoning traces generated by \texttt{Qwen2.5-14B} models fine-tuned on the original verbose data (top) and our ConMax-compressed data (bottom). Unlike the baseline, the model trained with ConMax generates a streamlined and logically coherent chain-of-thought, retaining essential reasoning patterns with significantly reduced token usage.}
    % \caption{Our model automatically detects and eliminates redundant verification steps and verbose phrasing found in the original CoT (top). The result is a streamlined, logically coherent CoT (bottom) that retains the underlying reasoning patterns.}
    \label{fig:intro}
    \vspace{-0.2cm}
\end{figure}
% A critical paradigm for evolving Large Language Models (LLMs) into LRMs involves supervised fine-tuning on datasets containing long reasoning traces, which are often distilled from existing strong LRMs (e.g., DeepSeek-R1~\citep{guo2025deepseek}). Under this paradigm, the quality of the training data is paramount: the complexity of the reasoning behaviors and the level of redundancy within the training data directly dictate the reasoning performance and inference efficiency of the resulting LRM. 
A pivotal paradigm in the pipeline of post-training Large Language Models (LLMs) into LRMs utilizes supervised fine-tuning (SFT) as a ``cold start'' mechanism. This process relies on datasets enriched with extensive reasoning traces, typically distilled from established, high-performance LRMs such as DeepSeek-R1~\citep{guo2025deepseek}. Within this framework, data quality is the decisive factor: the sophistication of the reasoning behaviors and the degree of redundancy in the cold-start training corpus directly govern both the reasoning capabilities and the inference efficiency of the resulting model. 
To address the issue of verbosity, recent studies tend to compress reasoning chains via heuristic rules, such as truncating paths to retain only the initial thinking blocks~\citep{DBLP:journals/corr/abs-2412-21187,qu2025optimizing} or removing steps based on perplexity~\citep{cui2025stepwise}. However, these heuristic approaches risk disrupting logical coherence and undermining the intricate reasoning pattern essential for complex tasks, while also depending heavily on brittle, tuned thresholds. While search-based compression~\citep{lu2025retro,wang2025r1} attempts to iteratively refine each reasoning step by invoking LLMs multiple times. However, this approach incurs prohibitively high sampling costs, yet fails to achieve significant length reduction empirically.

% Model-based copression has potential; and is not trival (explanation + prompt-based can not work emprically)
In this paper, we explore whether an LLM can automatically compress redundant reasoning chains by identifying and pruning superfluous components, outputting a condensed chain of thought without relying on complex heuristic metrics or step-level search procedures as discussed above. We distinguish this objective from abstractive summarization~\citep{stiennon2020learning} because the compressed chain must retain essential cognitive behaviors such as sub-goal decomposition and verification, to ensure its utility as training data to benefit downstream models. However, achieving this is non-trivial. First, our preliminary experiments reveal that simply applying prompt engineering to off-the-shelf LLMs is insufficient because models frequently conflate conciseness with simplification. Second, training a dedicated LLM for this task is challenging due to the scarcity of ground truth, specifically the absence of well-designed mappings from verbose reasoning chains to their optimal compressed forms.

% Our Method, rl, different confidence
%Our method. 
To address the above challenges, we propose \textbf { ConMax} (\textbf{Con}fidence-\textbf{Max}imizing Compression), which formulates the compression of reasoning traces as a reward-driven optimization problem. Without the need for human annotation, ConMax dynamically trains a compression policy $\pi_\theta$ to identify and prune unnecessary verbose steps while preserving intricate reasoning patterns.  Specifically, the compression policy is optimized by maximizing two confidence-based rewards derived from a frozen auxiliary LRM $\pi_\phi$: \textit{Answer Confidence}, which ensures predictive fidelity by verifying that the compressed trace retains sufficient causal logic to derive the correct answer, and \textit{Thinking Confidence}, which evaluates the reasoning content itself to ensure the compressed sequence remains linguistically coherent and logically valid.  Notably, instead of imposing explicit length penalties that may hinder exploration, we rely on a compression-oriented system prompt combined with this reward framework to guide the policy in intrinsically identifying and pruning linguistic redundancy while preserving the reasoning chain's integrity.
Through comprehensive experiments on five challenging reasoning datasets, we show that models fine-tuned on ConMax-processed data achieve a significantly better balance between accuracy and efficiency than those trained on raw or heuristically compressed data. Empirical results substantiate the efficacy of our framework: when fine-tuning \texttt{Qwen2.5-7B-Instruct}, our method reduces inference length by 43\% compared to strong baselines with a negligible accuracy drop of only 0.7\%. Additionally, ablation studies validate our confidence-based reward design, and further analysis confirms the robustness of these efficiency gains.
%, persisting even after further optimization via reinforcement learning.
%Through extensive experiments, we show that models fine-tuned on ConMax-processed datasets achieve a better balance between accuracy and efficiency compared to those trained on raw or heuristically compressed data. We conduct comprehensive experiments on five challenging reasoning datasets. Empirical results substantiate that our method significantly compresses reasoning paths without compromising performance, underscoring the efficacy of our framework. Notably, when fine-tuning \texttt{Qwen2.5-7B-Instruct} with CoT data generated by our method, the inference length is reduced by 43\% compared to a strong baseline, with a negligible average accuracy drop of only 0.7\%. Ablation studies further validate the rationale behind our internal confidence-based reward design. Moreover, our analysis indicates that this conciseness is robust, as models initialized with our data continue to exhibit superior efficiency even after further optimization with reinforcement learning.
% \end{itemize}
%\vspace{15cm}
% experiments: main results（performance and efficiency, good base for further RL)

% % Our method 
%\clearpage 
\section{Related Works}
\vspace{-2mm}
\subsection{Efficient Reasoning}
% TODO: OUR那里引用并介绍更多的工作，参考2025 EMNLP ConCISE
While Large Reasoning Models (LRMs) achieve superior performance by dedicating more time to thinking, this incurs significant computational costs. Empirical studies indicate that LRMs often engage in redundant ``overthinking''~\citep{DBLP:journals/corr/abs-2412-21187}, motivating research into three main categories of optimization methods. The first category focuses on adaptive reasoning, where models dynamically decide whether to activate a slow-thinking mode. Approaches like AdaCoT~\citep{lou2025adacot} and others~\citep{li2025tl} tune the ratio of thinking to non-thinking data during training based on question difficulty. In the RL phase, recent methods balance generation length by adjusting sampling distributions or using selective masking to prevent excessive token generation~\citep{wan2025adapthink,wang2025adaptive}. 
A second stream of research targets efficiency via length constraints and rewards in the RL phase. Approaches range from incentivizing brevity by scoring shorter rollouts higher~\citep{arora2025training,team2025kimi}, to enforcing structural bounds via iterative pruning~\citep{hou2025thinkprune} or constrained policy optimization~\citep{aggarwal2025l1}. Furthermore, methods such as AdaCtrl~\citep{huang2025adactrl} and LASER~\citep{liu2025learn} implement complexity-aware reward mechanisms, dynamically aligning token budgets with the difficulty of the query.~\citet{wu2025arm} propose Ada-GRPO, which utilizes a format diversity reward to mitigate format collapse, guiding the model to adaptively select efficient reasoning formats (e.g., short CoT or code).
The third category compresses reasoning chains to reduce token usage while maintaining performance. Several approaches truncate traces by identifying valid segments, either via forced early stopping~\citep{qu2025optimizing} or by retaining only the initial correct solutions~\citep{DBLP:journals/corr/abs-2412-21187}. \citet{qiao2025concise} introduces a confidence-guided framework actively suppressing redundant reflections via confidence injection and post-answer early stopping. Finally, search-based strategies are utilized to optimize compression, using either prompt-driven selection~\citep{wang2025r1} or iterative thought exploration akin to Monte Carlo Tree Search~\citep{lu2025retro}.
\subsection{Reinforcement Learning for Advanced Reasoning}
Reinforcement Learning with Verifiable Rewards (RLVR) has emerged as a pivotal paradigm for elevating the problem-solving capabilities~\citep{li2025system,guo2025deepseek,jin2025search} of LLMs, catalyzing the inception of powerful large reasoning models~\citep{jaech2024openai,guo2025deepseek,comanici2025gemini}. By grounding optimization in straightforward rule-based rewards, RLVR elicits LLMs'complex cognitive behaviors such as verification, backtracking, and subgoal setting~\citep{cen2025behavior,gandhi2025cognitive}.  Furthermore, recent studies indicate that even without external verifiable rewards, RL driven by intrinsic signals such as self-consistency~\citep{zuo2025ttrl}, token-level entropy~\citep{agarwal2025unreasonable}, self-certainty~\citep{zhao2025learning} and semantic entropy~\citep{zhang2025right} can effectively stimulate deep thinking. In this paper, we present a framework that harnesses multi-faceted confidence estimates as reward proxies, incentivizing LLMs to discover concise yet equally informative reasoning paths.

\vspace{-1mm}
\section{Methodology}
\vspace{-1mm}
\subsection{Problem Formulation}

We consider a standard System-2 reasoning setting involving a Large Reasoning Model (LRM), denoted as $\pi_\phi$. Given a query $x$, the LRM $\pi_\phi$ first generates a verbose reasoning trace $z$ (e.g., a chain-of-thought) and subsequently produces a final answer $y$. Formally, the generation process of the original model is factorized as $p_\phi(y, z | x) = p_\phi(y | x, z) p_\phi(z | x)$. We assume the original trace $z$ is sufficient for $\pi_\phi$ to derive the correct answer $y$.

\begin{figure*}
\centering
\includegraphics[width=0.99\textwidth]{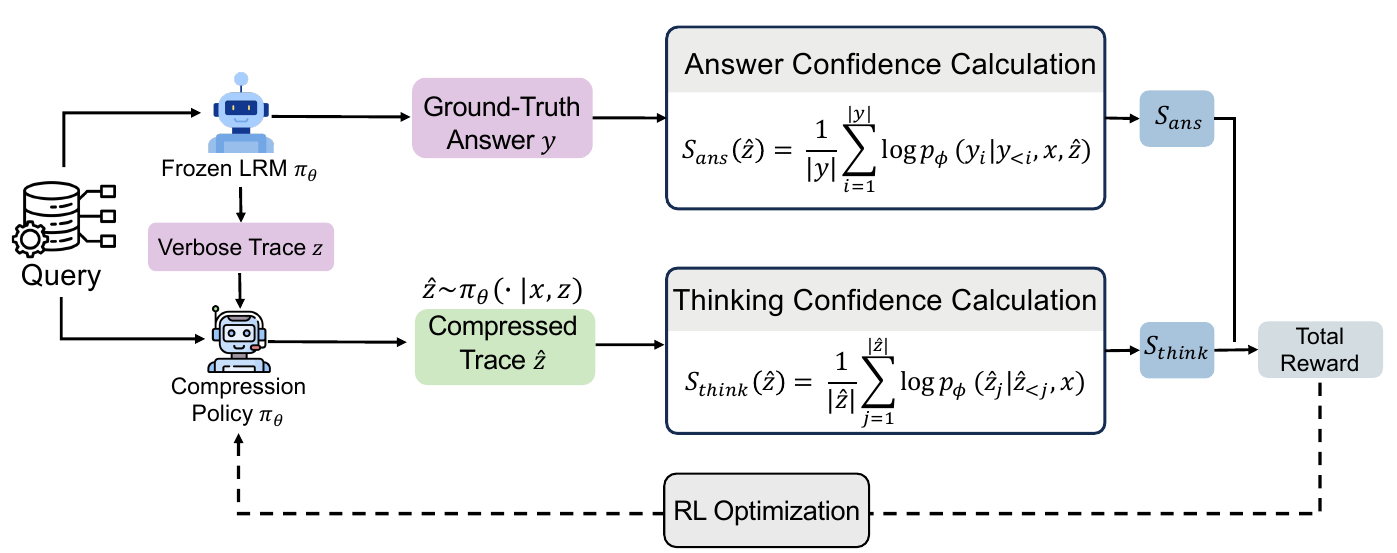}
\caption{Overview of ConMax. The framework uses reinforcement learning to train a policy $\pi_\theta$ for compressing originally verbose reasoning chains. A composite reward function guides the training by balancing Answer Confidence and Thinking Confidence during compression.}
\vspace{0.2cm}
\label{fig:methodology}
\end{figure*}

Our goal is to learn a compression policy, parameterized by $\theta$ and denoted as $\pi_\theta$, which distills the verbose trace $z$ into a concise version $\hat{z}$. The compression policy takes the system prompt $p$, the query $x$, and the original trace $z$ as input: $\hat{z} \sim \pi_\theta(\cdot | p, x, z)$. Here, $p$ serves as the instruction guiding the policy to perform the compression task. For brevity, we omit $p$ in subsequent notations and denote the conditional probability simply as $\pi_\theta(\cdot | x, z)$. The specific content of the system prompt is provided in Figure~\ref{fig:system_prompt}. Ideally, the compressed trace $\hat{z}$ should satisfy two criteria:
\begin{enumerate}
\item \textbf{Predictive Fidelity:} The compressed trace $\hat{z}$ must retain critical information sufficient to lead to the correct answer $y$ with high probability.
\item \textbf{Reasoning Validity:} The compressed trace $\hat{z}$ should remain a logically valid and coherent reasoning path, maintaining the internal consistency of the original thought process.
\end{enumerate}
\subsection{Confidence-Maximizing Compression}

To train the compression policy $\pi_\theta$ without requiring expensive human annotations, we propose a reward-based framework driven by two confidence scores derived from the frozen LRM: Answer Confidence ($S_{\text{ans}}$) and Thinking Confidence ($S_{\text{think}}$).

\paragraph{Answer Confidence}
This score measures the predictive fidelity of the compressed trace. It is calculated by querying the frozen LRM $\pi_\phi$ with the compressed trace $\hat{z}$ to see how well it predicts the ground-truth answer $y$. We define $S_{\text{ans}}$ as the length-normalized log-probability of $y$ under $\pi_\phi$:
\begin{equation}
S_{\text{ans}}(\hat{z}) = \frac{1}{|y|} \sum_{i=1}^{|y|} \log p_\phi(y_i \mid y_{<i}, x, \hat{z}).
\end{equation}
A high $S_{\text{ans}}$ indicates that $\hat{z}$ successfully captures the critical reasoning logic required by the original model $\pi_\phi$ to solve the problem.

\paragraph{Thinking Confidence}
This score ensures that the compressed trace maintains semantic coherence and validity according to the strong teacher model. It is defined as the average log-probability of the compressed trace $\hat{z}$ as evaluated by the frozen LRM $\pi_\phi$:
\begin{equation}
S_{\text{think}}(\hat{z}) = \frac{1}{|\hat{z}|} \sum_{j=1}^{|\hat{z}|} \log p_\phi(\hat{z}_j \mid \hat{z}_{<j}, x).
\end{equation}
By maximizing $S_{\text{think}}$, we encourage the compression policy to generate traces that follow a distribution likely to be produced by the more capable LRM, preventing the generation of disjointed keywords or broken syntax.

\paragraph{Total Reward}
The final confidence-based reward $R_{c}(\hat{z})$ for a generated compressed trace is the sum of these two confidence scores:
\begin{equation}\label{eq:total_reward.}
R_c(\hat{z}) = S_{\text{ans}}(\hat{z}) + \beta \cdot S_{\text{think}}(\hat{z}),
\end{equation}
where $\beta$ is a hyperparameter balancing the importance of answer accuracy versus reasoning coherence.

\subsection{Reinforcement Learning Framework}

To optimize the policy model $\pi_\theta$, we adopt the Group Relative Policy Optimization (GRPO)~\citep{shao2024deepseekmath} algorithm. Unlike  PPO~\citep{schulman2017proximal} which requires a separate value function critic, GRPO estimates the baseline using the group average of rewards, reducing training overhead.

Specifically, for each training instance consisting of a query $x$ and a verbose trace $z$, we first use the current policy $\pi_\theta$ to sample a group of $G$ candidate compressed traces $\{\hat{z}_1, \hat{z}_2, \dots, \hat{z}_G\}$. The advantage $A_{i,t}$ for the $t$-th token of  $i$-th candidate $\hat{z}_i$ is calculated by normalizing its reward with the mean and standard deviation of rewards within the group:
\begin{equation}
A_{i,t} = \frac{R_c(\hat{z}_i) - \text{mean}(\{R_c(\hat{z}_1), \dots, R_c(\hat{z}_G)\})}{\text{std}(\{R_c(\hat{z}_1), \dots, R_c(\hat{z}_G)\})}.
\end{equation}

The GRPO training objective is then formulated to maximize the advantage while constraining the policy update using a KL divergence penalty:

\begin{align}
\mathcal{J}&(\theta) = \mathbb{E}_{x \sim \mathcal{D}, \{\hat{z}_i\}_{i=1}^G \sim \pi_{\theta_{\text{old}}}(\cdot|x)}\frac{1}{G} \sum_{i=1}^G \frac{1}{|\hat{z}_i|} \sum_{t=1}^{|\hat{z}_i|} \nonumber \\
& \left[ \min \left( r_{i,t} A_{i,t}, \text{clip}(r_{i,t}, \epsilon) A_{i,t} \right) - \beta_{\text{KL}} \mathbb{D}_{\text{KL}} \right],
\end{align}

where $r_i = \frac{\pi_\theta(\hat{z}_{i,t}| x, z;\hat{z}_{i,<t})}{\pi_{\theta_{\text{old}}}(\hat{z}_{i,t}| x, z;\hat{z}_{i,<t})}$ denotes the importance sampling ratio, $\text{clip}(r_{i,t}, \epsilon)$ denotes the clipping function $\text{clip}(r_{i,t}, 1-\epsilon, 1 + \epsilon)$, and $\beta_{\text{KL}}$ regulates the divergence $ \mathbb{D}_{\text{KL}}$ between the trained policy $\pi_\theta$ and the reference model $\pi_{\text{ref}}$. 
%\zx{}

%However, directly summing $R_{\text{info}}$ and $R_{\text{len}}$ is problematic due to scale discrepancies: $R_{\text{info}}$ typically consists of log-probabilities (negative values), while $R_{\text{len}}$ is a ratio typically between 0 and 1. To address this, we apply group-based normalization (whitening) to each reward component independently during the rollout phase. The normalized rewards, $\tilde{R}_{\text{info}}$ and $\tilde{R}_{\text{len}}$, are then combined using a weighting factor $\lambda$:
%\begin{equation}
%    R(\hat{z}) = \tilde{R}_{\text{info}}(\hat{z}) + \lambda \cdot %\tilde{R}_{\text{len}}(\hat{z}),
%\end{equation}
%where $\lambda$ modulates the trade-off between the informativeness of the trace and its conciseness. The policy $\pi_\theta$ is updated to maximize this total expected reward.
\section{Experiments}

\subsection{Policy Evaluation Procedure}
% 描述：通过对 R1-7B Rejct sampling 得到一批原始数据。取其中一半做为Policy 的训练。训练完后取另外一半过 policy，得到压缩思维链；拉起SFT以比较性能。
We randomly sample 20k questions from the NuminaMath~\citep{li2024numinamath} subset of the OpenThoughts-114k\footnote{https://huggingface.co/datasets/open-thoughts/OpenThoughts-114k} dataset. For each question, we perform rejection sampling using R1-7B, retaining only those responses whose answers are equivalent to the ground truth. This process yields a dataset $\mathcal{D} = \{(x, z, y)\}$, where both the reasoning chain $z$ and the final answer $y$ for each data point are generated by R1-7B. We randomly partition $\mathcal{D}$ into two disjoint subsets: $\mathcal{D}_{\mathrm{sft}}$, containing about 9k samples, and $\mathcal{D}_{\mathrm{rl}}$, comprising about 6.5k samples. The subset $\mathcal{D}_{\mathrm{rl}}$ is utilized to train a policy $\pi_\theta$ for compressing reasoning chains.

Upon completion of policy training, each $(x, z)$ pair in $\mathcal{D}_{\mathrm{sft}}$ is processed by $\pi_\theta$ to produce a compressed reasoning chain $\hat{z}$, resulting in a set of compressed samples $(x, \hat{z})$. Subsequently, R1-Distill-Qwen-7B is employed to generate completions for text sequences of the form ``$x \circ \hat{z}$,'' thus constructing the final compressed dataset $\widehat{\mathcal{D}}_{\mathrm{sft}}$.  Ideally, an effective policy $\pi_\theta$ is expected to produce a compressed dataset $\widehat{D}_{\mathrm{sft}}$ such that models fine-tuned on $\widehat{D}_{\mathrm{sft}}$ achieve performance comparable to those fine-tuned on $D_{\mathrm{sft}}$, while yielding a substantial reduction in token consumption.

\subsection{Experiment Setup}
\paragraph{Implementation Details}
%\zx{[complete this]}
The compression policy is initialized from DeepSeek-R1-Distill-Qwen-7B~\footnote{https://huggingface.co/deepseek-ai/DeepSeek-R1-Distill-Qwen-7B} and trained using the Verl~\citep{sheng2025hybridflow} framework. The training process spans 206 steps, with a batch size configuration of 32 queries and 8 rollouts per query. We utilize the Adam optimizer with a learning rate of $1\text{e-}6$. During rollouts, we employ a sampling temperature of 1.0, top-k of -1, and top-p of 1, with both prompt and response lengths capped at 12,888 tokens. Additionally, we apply KL regularization with $\beta_{\text{KL}}=0.001$ and set the entropy coefficient to 0. For the SFT stage, we use LLaMAFactory~\citep{zheng2024llamafactory}.All models are fine-tuned for 6 epochs using a learning rate of $5\text{e-}6$, a batch size of 64, and a sequence length of 16K. We employ a cosine learning rate scheduler with no warmup.

\paragraph{Models and Baselines}
We conduct supervised fine-tuning on top of three Instruct models: Qwen2.5-\{3B/7B/14B\}-Instruct~\footnote{https://huggingface.co/Qwen/Qwen2.5-\{3B/7B/14B\}-Instruct} to comprehensively evaluate the effectiveness of the compressed reasoning data. To validate the efficacy of our method, we compare our proposed compression strategy against two baselines: \textit{Original Responses}, where we fine-tune directly on the original dataset $D_{\text{sft}}$ using unprocessed reasoning chains, and \textit{Prompting-Based Compression}, which employs prompting without reinforcement learning to guide the policy model in compressing the reasoning trajectories.

\paragraph{Evaluation Protocol}
To assess the efficacy of our method, we conducted evaluations across five demanding reasoning benchmarks: AIME2025, MATH500~\citep{hendrycks2021measuring}, AMC23, MINERVA, and GPQA~\citep{rein2024gpqa}. For inference, we set the maximum generation length to 32,768 tokens, utilizing a sampling temperature of $0.6$, $\text{top-p}=0.95$, and $\text{top-k}=-1$. Given the limited size of the AIME2025 and AMC23, we conduct five independent runs and report the average performance. Reported token counts of the generated responses are determined using the specific tokenizer associated with the language model. 
Finally, answer verification was performed using a rule-based approach via the \texttt{Math-Verify}\footnote{https://github.com/huggingface/Math-Verify} library.

% \paragraph{Baselines.}
% We compare our method against several strong baselines for generating compressed reasoning datasets, each representing a different strategy for dataset creation and subsequent model fine-tuning. \textbf{Original Responses} serves as the primary baseline, where we fine-tune directly on the original dataset $D_{\text{sft}}$ using the unprocessed reasoning chains. \textbf{Shortest Correct Responses} utilizes a heuristic to select concise, correct reasoning paths: for each question in $D_{\text{sft}}$, we sample eight responses that reach the correct answer and select the shortest one by token count, yielding $D_{\text{shortest}}$, a dataset of the shortest correct reasoning responses for each question. 
% %\textbf{Truncated Reasoning Trajectory}, following \citep{DBLP:journals/corr/abs-2412-21187}, retains only the first two complete ``thinking blocks'' that both lead to the correct answer for each reasoning trajectory in $D_{\text{sft}}$. 
% \textbf{Prompting-Based Compression} employs prompting---without reinforcement learning---to guide the policy model in compressing the reasoning trajectories of samples from $D_{\text{sft}}$, resulting in the final prompting-based compressed dataset. For all datasets obtained from these methods, as well as the dataset compressed by our proposed method, we conduct supervised fine-tuning on top of two Instruct models, Qwen2.5-7B/14B-Instruct\footnote{https://huggingface.co/collections/Qwen/qwen25-66e81a666513e518adb90d9e}, to evaluate the effectiveness of each dataset.

\begin{table*}[!h]
    \small
    \renewcommand{\arraystretch}{1.22} 
    \centering
    \setlength{\tabcolsep}{1.4pt} 
    
    \begin{tabular*}{\textwidth}{@{\extracolsep{\fill}} l cccccc cccccc }
        \toprule
        \multirow{2}{*}{\textbf{Method}} & \multicolumn{6}{c}{\textbf{Accuracy~\%}~$\uparrow$} & \multicolumn{6}{c}{\textbf{Token Length}~$\downarrow$} \\
        \cmidrule(lr){2-7} \cmidrule(lr){8-13} 
         & AIME & MATH & AMC & Min. & GPQA & \textit{Avg. (Perf. Loss)} & AIME & MATH & AMC & Min. & GPQA & \textit{Avg. (Comp. Rate\%)} \\
        \midrule \midrule
        Qwen2.5-3B   & 1.3 & 64.8 & 38.0 & 28.3 & 30.8 & 32.6 & 939 & 593 & 958 & 629 & 630 & 750 \\
        \hdashline % <--- 这里加了虚线
        Original & 9.3 & 69.6 & 43.0 & 25.4 & 35.0 & 36.5 & 9,256 & 4,255 & 7,852 & 4,603 & 4,408 & 6,075 \\
        Prompt-based & 1.3 & 55.8 & 27.0 & 23.2 & 35.4 & 28.5 ($\down{8.0}$) & 1,186 & 1,057 & 1,101 & 1,225 & 988 & 1,111 ($\up{81.7\%}$) \\
        %Ours (wo. coh) & 11.3 & 70.0 & - & - & - & - & - & 2485 & - & - & - & - \\
        Ours & 9.3 & 67.4 & 48.5 & 29.4 & 41.4 & 39.2 ($\up{2.7}$) & 6,390 & 3,357 & 4,686 & 3,512 & 3,247 & 4,238 ($\up{30.2\%}$)\\
        \midrule
        Qwen2.5-7B      & 8.0  & 75.8 & 51.9 & 39.0 & 38.9 & 45.1 & 926   & 615  & 915  & 661  & 704  & 765 \\ 
        \hdashline % <--- 这里加了虚线
        Original        & 22.7 & 84.8 & 71.5 & 42.1 & 50.0 & 54.2  & 14,019 & 5,247 & 9,417 & 6,362 & 7,968 & 8,603 \\
        Prompt-based    & 2.7  & 68.6 & 39.0 & 37.1 & 42.9 & 38.1 ($\down{16.1}$) & 1,146  & 1,034 & 1,155 & 1,017 & 829  & 1,036 ($\up{88.0\%}$) \\
        %Shortest        & 21.3 & 83.0 & 64.5 & 43.0 & 50.0 & 52.4 & 10,000 & 3,613 & 6,841 & 4,637 & 5,359 & 6,070 \\
        %Ours (wo. coh) & 18.7 & 82.2 & 62.8 & 41.7 & 49.6 & 51.4 ($\down{2.8}$) & 4,926 & 3,206 & 5,088 & 3,839 & 4,261 & 4,264 ($\up{50.4\%}$) \\
        Ours & 24.7 & 81.8 & 67.7 & 41.9 & 51.5 & 53.5 ($\down{0.7}$) & 7,062 & 3,691 & 5,224 & 4,039 & 4,517 & 4,906 ($\up{43.0\%}$) \\
        %Ours (pf0p01cotr1p2) & 21.3 & 83.4 & 65.8 & 41.2 & 49.0 & 52.1 & 7,226 & 3,474 & 5,930 & 3,610 & 4,162 & 4,880 \\
        %Ours (pf0p01cotr1p2ep5) & 22.0 & 81.6 & 67.0 & 41.3 & 46.4 & 51.6 & - & - & - & - & - & - \\
        %Ours (pf0cotr1) & 21.3 & 83.4 & 67.0 & 43.0 & 47.2 & 52.4 & - & - & - & - & - & - \\
        %Ours (pf0cotr2) & 25.3 & 81.2 & 66.0 & 41.9 & 46.4 & 52.2 & - & - & - & - & - & - \\ 
        %Ours (pf0cotr0p8) & 22.0 & 81.8 & 64.8 & 46.3 & 48.7 & 52.7 & - & - & - & - & - & - \\ 
        \midrule
        Qwen2.5-14B     & 12.7 & 79.6 & 61.0 & 43.8 & 46.0 & 48.6 & 943   & 580  & 890  & 612  & 716  & 748 \\ 
        \hdashline % <--- 这里加了虚线
        Original        & 31.3 & 88.8 & 79.3 & 47.8 & 58.6 & 61.1 & 11,236 & 3,877 & 6,733 & 4,822 & 5,735 & 6,481 \\
        Prompt-based    & 6.7  & 75.2 & 50.0 & 44.1 & 41.9 & 43.6 ($\down{17.5}$) & 1,141  & 1,006 & 1,179 & 1,138 & 885  & 1,070 ($\up{83.5\%}$) \\
        %Ours (wo. coh) & 21.3 & 87.8 & 73.5 & 50.0 & 57.1 & 56.9 ($\down{4.2}$) & 5304 & 2858 & 3991 & 3259 & 3138 & 3,710 ($\up{42.8\%}$) \\
        %Shortest        & 32.0 & 90.2 & 80.0 & 50.7 & 62.6 & 63.1 & 9,831  & 3,034 & 5,655  & 3,615  & 5,177  & \textbf{5,462} \\
        %Shortest (valid)  & 32.0 & 90.2 & 80.0 & 50.7 & 62.6 & 63.1 & 6028  & 2612 & 4345 & 2965 & 4683 & 3651 \\
        %Ours (pf0cotr1p2) & 27.3 & 87.2 & 77.3 & 47.8 & 58.1 & 59.5 & 10,645 & 4,012 & 6,212 & 4,022 & 4,437 & 5,866 \\
        %Ours (pf0p01cotr1p2) & 29.3 & 88.4 & 73.5 & 48.5 & 57.4 & 59.4 & 7,882 & 3054 & 5,796 & 3,608 & 4,806 & 5,029 \\
        %Ours (pf0p01cotr1p2ep5) & 28.0 & 87.2 & 73.0 & 47.8 & 62.6 & 59.7 & 8758 & 3305 & 5703 & 3466 & 3401 & 4927 \\
        %Ours (pf0p01cotr1p2, valid) & 29.3 & 88.4 & 73.5 & 48.5 & 57.4 & 59.4 & 5174 & 2645 & 4937  & 3343 & 4626 & 4145 \\
        Ours  & 30.0 & 87.2 & 78.9 & 47.6 & 54.5 & 59.6 ($\down{1.5}$) & 7,688 & 3,045 & 4,891 & 3,574 & 3,765 & 4,593 ($\up{29.1\%}$)\\
        %Ours (pf0cotr1p2ep5, valid) & 30.0 & 87.2 & 78.9 & 47.6 & 54.5 & 59.6 & 4743 & 2582 & 4189 & 2988 & 3758 & 3652 \\
        \bottomrule
    \end{tabular*}
    \caption{Main experiment results on \texttt{Qwen-2.5-Instruct} model family. Min. denotes MINERVA for short.}
    \label{tab:main_results.}
    \vspace{0.3cm}
\end{table*}
%\vspace{-0.8cm}

% \subsection{Analysis on Main Results}
% Table~\ref{tab:main_results.} shows the performance of our ConMax method and all baselines on \texttt{Qwen-2.5} model family 
% %\md{[analysis for the main result table]}
% As we can see, ConMax brings consistent improvements in token length, while sacrificing marginal performance in both complex mathematical reasoning challenges like AIME25 and science QA benchmarks such as GPQA. In general, our method brings an average performance decrease of less than 0.7 points and drastically compresses the chain-of-thought up to 43.0\%, showcasing the efficacy and robustness of our method. We further analyze such a performance-efficiency tradeoff as follows.

\subsection{Analysis of Main Results}
Table~\ref{tab:main_results.} presents the performance of our ConMax method compared to all baselines across the \texttt{Qwen-2.5} model family. 
%\md{[analysis for the main result table]}
As illustrated, ConMax delivers consistent reductions in token length while sacrificing only marginal performance on both complex mathematical reasoning challenges (e.g., AIME25) and science QA benchmarks (e.g., GPQA). Overall, our method entails an average performance decrease of less than 1.5 points while significantly compressing CoT length by up to 43.0\%, demonstrating both efficacy and robustness. We further analyze this trade-off between performance and efficiency below.

\paragraph{Performance Loss}
%Regarding the performance loss that ConMax compression brings, it is maintained as the accepted level of 1.5 points loss in \texttt{Qwen-2.5-14B} and 0.7 in \texttt{Qwen-2.5-17B}. For hard mathematical problems, ConMax maintains high fidelity and keep the loss in AIME2025 less than 1.5 in \texttt{Qwen-2.5-14B} and even brings an improvement of 2 points in \texttt{Qwen-2.5-7B}. For easier mathematical problems like MATH and AMC, ConMax can achieve preserving less than 1 point loss in \texttt{Qwen-2.5-14B}. For GPQA, ConMax even brings performance boost of 1.5 points in \texttt{Qwen-2.5-7B}.
Regarding the impact of ConMax compression on performance, the degradation remains within an acceptable margin, specifically a 1.5-point loss for \texttt{Qwen-2.5-14B} and 0.7 points for \texttt{Qwen-2.5-7B}. On challenging mathematical benchmarks such as AIME2025, ConMax maintains high fidelity, limiting the loss to under 1.5 points for \texttt{Qwen-2.5-14B} while yielding a 2-point improvement for \texttt{Qwen-2.5-7B}. For other mathematical benchmarks like AMC and Min, ConMax preserves performance effectively, with the 14B model exhibiting negligible degradation (e.g., less than 0.5 points). Notably, on the scientific benchmark GPQA, ConMax demonstrates unexpected gains, improving the \texttt{Qwen-2.5-7B} score by 1.5 points and achieving an even larger increase of 6.4 points for the 3B model.
%In addition, comparing different sizes of \texttt{Qwen-2.5}, it is noteworthy that the performance loss gradually increases as the size increases from 3B to 14B. This implies different foundational capabilities of models, as larger models are more capable of understanding more fine-grained patterns from the Original, part of which our method ConMax inevitably discards. On the contrary, these traces hinder small models' reasoning; instead, given the results, there is even an average performance uplift of 2.7 when applying the ConMax-revised dataset.
Furthermore, a comparison across different \texttt{Qwen-2.5} model sizes reveals that performance loss gradually increases as the model size scales from 3B to 14B. This trend suggests a divergence in foundational capabilities: larger models are better equipped to interpret fine-grained patterns in the original data, some of which are inevitably discarded by ConMax. Conversely, these intricate traces appear to hinder the reasoning of smaller models. Consequently, applying the ConMax-revised dataset to smaller models results in an average performance uplift of 2.7 points.

%In addition, we visualize the distribution of generated token lengths for correctly predicted cases, comparing ConMax against the Original Baseline in Figure~\ref{fig:token_length_distribution.}. As shown in the figure, the ConMax distribution exhibits a pronounced peak at shorter token lengths, indicating that it generates more concise reasoning compared to the baseline.

%\md{[analysis for performance loss]}
\paragraph{Compression Rate}
% 3B和7B的压缩率维持在30%左右， 14B在43%，ConMax-finetune过的Qwen 7B family token数量维持在5000token以下，平均每题节约2000 token，这样的实验结果充分验证了ConMax的效率和有效性。另外，和Prompt-based的对比我们可以看出虽然没有经过RL对齐过的optimizer可以很大程度上的压缩cot长度，但是它在7B和14B模型上也有超过百分之15的平均性能损失，其表现甚至弱于Qwen 2.5 Instruct版本的模型。这证明了模型本身对如何保留对sft有用的reasoning traces没有相关的知识，而ConMax设计的RL阶段让optimizer关于如何修改cot的有效的知识和信息。相比于Prompt-based激进而且对性能有很大伤害的修改，ConMax在压缩和保持性能之间做到了很好的平衡，with 损失可控的情况下也大幅缩短了cot token length.
%\md{[analysis for compression rate]}
The compression rate for the 3B and 14B models is approximately 30\%, while the 7B model achieves a rate of 43\%. Notably, the ConMax-finetuned \texttt{Qwen-2.5} family consistently keeps the token count below 5,000, saving an average of 2,000 tokens per problem. These results strongly validate the efficiency and effectiveness of ConMax. Furthermore, a comparison with prompt-based methods reveals that while methods lacking RL alignment  significantly reduce CoT length, they incur an average performance penalty exceeding 16\% on the 7B and 14B models, performing even worse than the standard \texttt{Qwen-2.5-Instruct} versions. This demonstrates that the model inherently lacks the knowledge required to identify and preserve reasoning traces beneficial for cold-start SFT. In contrast, ConMax's RL stage equips the optimizer with the necessary insights to modify the CoT effectively. Unlike the aggressive and detrimental revisions of prompt-based approaches, ConMax strikes a superior balance between compression and performance, significantly reducing CoT token length while keeping performance loss controllable.

\subsection{Accuracy Decomposition across Generated Token Length}
%\subsection{Effects of ConMax on following Mid-training}
%我们将Original Baseline和ConMax在\texttt{Qwen-2.5-7B}各个benchmark中的表现列在了Table~\ref{tab}中。我们分别统计了generation token少于4000（4k）， 8000（8k）和12000（12k）情况下的accuracy。From the result, we can clearly see that ConMax has endowed the LRM with significantly more effcient reasoning, with over 10 point improvement in Acc@4k on all benchmarks, and almost 在acc@4k/8k/12k具有压倒性优势。While Original has a slight advantage in overall accuracy, that is at the cost of using significantly more tokens to achieve it. On comparison, ConMax shows impressive reasoning efficiency, 也让LRM实际处理复杂问题的成本骤降。
%In addition, we have also drawn the distribution plot of f generated token lengths forcorrectly predicted cases, comparing ConMax against the Original Baseline in Figure~\ref{fig}. According to the figure, the ConMax distribution exhibits a stronger peak at shorter token lengths, indicating more concise generation compared to the baseline.
Table~\ref{tab:acc_at_token_limit.} details the performance of both the Original Baseline and ConMax using \texttt{Qwen-2.5-7B} across various benchmarks. We decompose the accuracy based on generated output lengths, specifically reporting results for token counts fewer than 4,000 (4k), 8,000 (8k), and 12,000 (12k).

\begin{table*}[htbp]
    %\small
    \centering
    \setlength{\tabcolsep}{8pt} 
    \resizebox{\textwidth}{!}{% 自动缩放表格以适应页面宽度
    \begin{tabular}{lcccccccccc}
        \toprule
        \multirow{2}{*}{\textbf{Dataset}} & \multicolumn{5}{c}{\textbf{ConMax ($\beta = 1.2$)}} & \multicolumn{5}{c}{\textbf{Original}} \\
        \cmidrule(lr){2-6} \cmidrule(lr){7-11}
        & \textbf{Acc (\%)~$\uparrow$} & \textbf{Avg Tok~$\downarrow$} & \textbf{@4k} & \textbf{@8k} & \textbf{@12k} & \textbf{Acc (\%)~$\uparrow$} & \textbf{Avg Tok~$\downarrow$} & \textbf{@4k} & \textbf{@8k} & \textbf{@12k} \\
        \midrule
        
        AIME 2025 & \textbf{24.7} & 7,062 & \textbf{0.12} & \textbf{0.22} & \textbf{0.25} & 22.7 & 14,019 & 0.05 & 0.18 & 0.22 \\
        %\textit{(Total/Corr)} & \textit{(150/37)} & & & & & \textit{(147/34)} & & & & \\
        \addlinespace
        
        MATH & 81.8 & \textbf{3,691} & \textbf{0.68} & \textbf{0.80} & \textbf{0.80} & \textbf{84.8} & 5,247 & 0.58 & 0.76 & \textbf{0.80} \\
        %\textit{(Total/Corr)} & \textit{(500/409)} & & & & & \textit{(500/424)} & & & & \\
        \addlinespace
        
        AMC & 67.7 & \textbf{5,224} & \textbf{0.46} & \textbf{0.61} & \textbf{0.66} & \textbf{71.5} & 9,417 & 0.34 & 0.47 & 0.59 \\
        %\textit{(Total/Corr)} & \textit{(198/134)} & & & & & \textit{(200/143)} & & & & \\
        \addlinespace
        
        MINERVA & 41.9 & \textbf{4,039} & \textbf{0.33} & \textbf{0.40} & \textbf{0.42} & \textbf{42.1} & 6,362 & 0.24 & 0.36 & 0.39 \\
        %\textit{(Total/Corr)} & \textit{(272/114)} & & & & & \textit{(272/115)} & & & & \\
        \addlinespace
        
        GPQA & \textbf{51.5} & \textbf{4,517} & \textbf{0.30} & \textbf{0.45} & \textbf{0.49} & 50.0 & 7,968 & 0.13 & 0.31 & 0.42 \\
        %\textit{(Total/Corr)} & \textit{(198/102)} & & & & & \textit{(197/99)} & & & & \\
        
        \bottomrule
    \end{tabular}%
    }
    \\[5pt]
    \footnotesize \textit{Note: Acc = Accuracy, Avg Tok = Average Tokens.}
    \caption{Accuracy of ConMax ($\beta = 1.2$) vs. Original results by generated token count across datasets. The best values are \textbf{bolded}.}
    \label{tab:acc_at_token_limit.}
\end{table*}
%\vspace{-0.8cm}
% Notably, ConMax achieves an improvement of over 10 percentage points in Acc@4k across all benchmarks, maintaining a dominant advantage across the Acc@4k, Acc@8k, and Acc@12k metrics.
The results demonstrate that ConMax significantly enhances the reasoning efficiency of the LRM.  Notably, ConMax achieves substantial improvements in Acc@4k across all benchmarks, averaging a gain of approximately 10 percentage points compared to the baseline. It maintains a dominant advantage in efficiency across the Acc@4k, Acc@8k, and Acc@12k metrics. Although the Original Baseline retains a marginal advantage in overall performance, this comes at the expense of significantly higher token consumption. In contrast, ConMax exhibits impressive reasoning efficiency, drastically reducing the inference cost required for LRMs to solve complex problems.

\begin{figure}[h]
    \centering
    \includegraphics[width=0.99\linewidth]{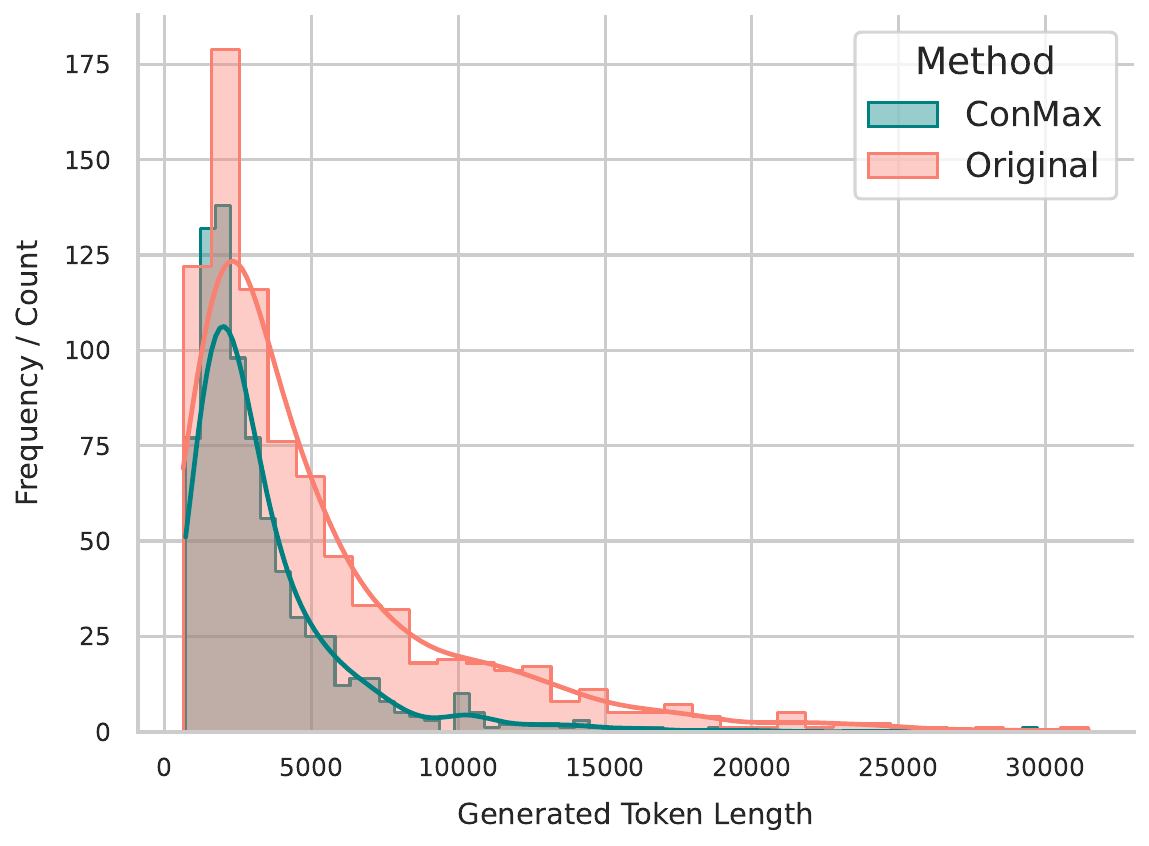}
    \caption{Distribution of generated token lengths for correctly predicted cases, comparing ConMax against the Original Baseline. The ConMax distribution exhibits a stronger peak at shorter token lengths, indicating more concise generation compared to the baseline.}
    \label{fig:token_length_distribution.}
    \vspace{-0.3cm}
\end{figure}

In addition, we visualize the distribution of generated token lengths for correctly predicted cases, comparing ConMax against the Original Baseline in Figure~\ref{fig:token_length_distribution.}. As shown in the figure, the ConMax distribution exhibits a pronounced peak at shorter token lengths, indicating that it generates more concise reasoning compared to the baseline.

\subsection{Design choices}
\subsubsection{Choice of $\beta$ in Thinking Confidence}
% Here we examine the effects of choosing different $\beta$ of $S_{think}(\hat{z})$ in our reward design in Eq.~\ref{eq:total_reward.}. In Table~\ref{tab:beta_choice.}, we can observe that higher $\beta$ from 0 to 1.2 can maintain more fidelity of the optimized CoT chain, preserving more information that helps LLMs to find the right solution with reasoning. When completely discarding $S_{think}(\hat{z})$ in the reward ($\beta = 0$), the optimizer will put the entire focus on how to generate a CoT that could easily derive the ground truth without considering whether this CoT can be easily generated by LRMs from scratch. Comprare with $\beta = 0.5$, our method can achieve 帕累托更优的result with less performnace loss and higher CoT compression rate. Nevertheless, excessively high $\beta$ value, such as 2.0, will also put too much attention on the model in generating CoT with high confidence frozen teacher model without enough focus on whether the corresponding CoT will derive correct answers.
In this section, we examine the impact of varying the coefficient $\beta$ for the $S_{think}(\hat{z})$ term within our reward formulation (Eq.~\ref{eq:total_reward.}). As shown in Table~\ref{tab:beta_choice.}, increasing $\beta$ from 0 to 1.2 improves the fidelity of the optimized CoT, preserving critical information that aids LLMs in reasoning toward the correct solution. When $S_{think}(\hat{z})$ is completely omitted ($\beta = 0$), the policy focuses exclusively on generating a CoT that derives the ground truth,
neglecting whether the generated chain remains linguistically coherent and follows the reasoning patterns.
% neglecting whether the LRMs can plausibly generate such a chain from scratch. 

Comparing different $\beta$ values reveals an interesting trade-off. While $\beta = 0.5$ achieves the highest compression rate (53.5\%), it comes at the cost of a notable drop in accuracy (2.3\%). In contrast, our chosen setting of $\beta = 1.2$ achieves an optimal balance: it drastically reduces the performance loss to a negligible 0.7\% while still maintaining a high compression rate of 43.0\%. However, an excessively high $\beta$ (e.g., 2.0) is detrimental. It overemphasizes alignment with the reasoning contents, distracting the model from deriving the correct final answer and leading to a performance drop of 2.0\%. Based on these results, we set $\beta = 1.2$ throughout all experiments in our work.
% Compared to $\beta = 0.5$, our chosen method achieves a Pareto-optimal result, balancing minimal performance loss with a higher CoT compression rate. However, an excessively high $\beta$ (e.g., 2.0) overemphasizes alignment with the frozen teacher model's confidence, thereby distracting the model from deriving the correct final answer. According to such results, we set $\beta = 1.2$ throughout all experiments in our work.
\begin{table}[!h]
    \small
    \renewcommand{\arraystretch}{1.22} 
    \centering
    \setlength{\tabcolsep}{0.1pt} 
    
    \begin{tabular*}{0.96\columnwidth}{@{\extracolsep{\fill}} lcccc }
        \toprule
        \multirow{1}{*}{\textbf{Method}} & \multicolumn{2}{c}{\textbf{Accuracy~\%}~$\uparrow$} & \multicolumn{2}{c}{\textbf{Token Length}~$\downarrow$} \\
        \cmidrule(lr){2-3} \cmidrule(lr){4-5} 
        & \textit{Avg.} & \textit{Perf. Loss} & \textit{Avg.} & \textit{Comp. Rate\%} \\
        \midrule \midrule
        %Qwen2.5-7B      & 8.0  & 75.8 & 51.9 & 39.0 & 38.9 & 45.1 & 926   & 615  & 915  & 661  & 704  & 765 \\ 
        Original    &    54.2  & - & 8,603 & - \\
        \hdashline % <--- 这里加了虚线
        %Prompt-based    & 2.7  & 68.6 & 39.0 & 37.1 & 42.9 & 38.1 ($\down{29.7\%}$) & 1,146  & 1,034 & 1,155 & 1,017 & 829  & 1,036 ($\up{88.0\%}$) \\
        %Shortest        & 21.3 & 83.0 & 64.5 & 43.0 & 50.0 & 52.4 & 10,000 & 3,613 & 6,841 & 4,637 & 5,359 & 6,070 \\
        %Ours (wo. coh) & 18.7 & 82.2 & 62.8 & 41.7 & 49.6 & 51.4 ($\down{5.2\%}$) & 4,926 & 3,206 & 5,088 & 3,839 & 4,261 & 4,264 ($\up{50.4\%}$) 
        %Ours ($\beta = 0.25$) & 21.0 & 83.4 & 65.8 & 41.2 & 49.0 & 52.1 & 7,226 & 3,474 & 5,930 & 3,610 & 4,162 & 4,880 \\
        Ours ($\beta = 0$) & 51.4 & $\down{2.8}$ & 4264 & $\up{50.4\%}$\\
        Ours ($\beta = 0.5$) & 51.9 & $\down{2.3}$ & 4001 & $\up{53.5\%}$\\
        Ours ($\beta = 0.8$) & 52.7 & $\down{1.5}$ & 5347 & $\up{37.8\%}$\\
        Ours ($\beta = 1.0$) & 52.4 & $\down{1.8}$ & 4811 & $\up{44.1\%}$\\
        Ours ($\beta = 1.2$) & 53.5 & $\down{0.7}$ & 4,906 & $\up{43.0\%}$\\
        Ours ($\beta = 2.0$) & 52.2 & $\down{2.0}$ & 5105 & $\up{40.7\%}$\\
        %Ours (pf0p01cotr1p2ep5) & 22.0 & 81.6 & 67.0 & 41.3 & 46.4 & 51.6 & - & - & - & - & - & - \\
        %Ours (pf0cotr1) & 21.3 & 83.4 & 67.0 & 43.0 & 47.2 & 52.4 & - & - & - & - & - & - \\
        %Ours (pf0cotr2) & 25.3 & 81.2 & 66.0 & 41.9 & 46.4 & 52.2 & - & - & - & - & - & - \\ 
        %Ours (pf0cotr0p8) & 22.0 & 81.8 & 64.8 & 46.3 & 48.7 & 52.7 & - & - & - & - & - & - \\ 
        \bottomrule
    \end{tabular*}
    \caption{Experiment results on choice of $\beta$ in \texttt{Qwen-2.5-7B-Instruct}.}
    % \vspace{-0.5cm}
    \label{tab:beta_choice.}
\end{table}

\subsubsection{Marginal Probability as Reward}
% Here we compare the effectiveness between separating the answer and thinking confidence and integrating them as marginal averaged log probability $R_{\mathrm{marginal}}$ of ground-truth $y$ and compressed trace $\hat{z}$ shown in Eq.~\ref{eq:margainal_prob.}. 
% \begin{equation}\label{eq:margainal_prob.}
% R_{\text{marginal}}(\hat{z}) = \frac{1}{|y \cup \hat{z}|} \sum_{i=1}^{|y \cup \hat{z}|} \log p_\phi([y \cup \hat{z}]_i \mid x).
% \end{equation}
% As shown in Table~\ref{tab:marginal_prob.}, ConMax with $R_{info}$ surpass $R_{marginal}$ in both performance retainment and compression rate. While $R_{\mathrm{marginal}}$ is the unbiased estimator of LRMs' sequence generation, $R_{\mathrm{marginal}}$ is heavily biased toward compressed trace generation, as its length is usually over a few thousand tokens and answer is less that one hundred, this will bury the generation confidence of $y$ in the $R_{\mathrm{marginal}}$  reward design. Thus, separating $S_{\mathrm{ans}}$ and $S_{\mathrm{think}}$ in can much more effectively ensure and constrain compressed CoT actually lead to the ground-truths. The result also strongly support our design, as $R_{\mathrm{marginal}}$ leads to more performance loss and no compression effect compared with $R_{\mathrm{marginal}}$.
In this section, we compare the effectiveness of separating answer and thinking confidence against integrating them via the marginal averaged log probability, denoted as $R_{\mathrm{marginal}}$. This metric, defined in Eq.~\ref{eq:margainal_prob}, combines the ground-truth $y$ and the compressed trace $\hat{z}$:
% \begin{equation}\label{eq:margainal_prob}
% R_{\text{marginal}}(\hat{z}) = \frac{1}{|y \cup \hat{z}|} \sum_{i=1}^{|y \cup \hat{z}|} \log p_\phi([y \cup \hat{z}]_i \mid x).
% \end{equation}
\begin{equation}\label{eq:margainal_prob}
R_{\text{marginal}}(\hat{z}) = \frac{1}{|y \circ \hat{z}|} \sum\limits_{i=1}^{|y \circ \hat{z}|} \log p_\phi([y \circ \hat{z}]_i \mid x).
\end{equation}
As shown in Table~\ref{tab:marginal_prob.}, ConMax utilizing $R_{{\mathrm{c}}}$ surpasses $R_{\mathrm{marginal}}$ in both performance retention and compression rate. While $R_{\mathrm{marginal}}$ serves as an unbiased estimator for sequence generation in LRMs, it becomes heavily biased toward the compressed trace during generation. Since the trace length often exceeds several thousand tokens while the answer is typically fewer than one hundred, the ground-truth reward $y$ is buried within the $R_{\mathrm{marginal}}$ reward signal. Consequently, separating $S_{\mathrm{ans}}$ and $S_{\mathrm{think}}$ is significantly more effective at ensuring that the compressed CoT is both accurate and concise. The results strongly support our design choice, as relying on $R_{\mathrm{marginal}}$ results in greater performance loss and leads to token length expansion rather than compression.
\begin{table}[!h]
    \small
    \renewcommand{\arraystretch}{1.22} 
    \centering
    \setlength{\tabcolsep}{0.1pt} % 稍微增加了列间距，因为列数变少了，这样看起来更宽敞
    
    \begin{tabular*}{0.96\columnwidth}{@{\extracolsep{\fill}} l cc cc }
        \toprule
        \multirow{1}{*}{\textbf{Method}} & \multicolumn{2}{c}{\textbf{Accuracy~\%}~$\uparrow$} & \multicolumn{2}{c}{\textbf{Token Length}~$\downarrow$} \\
        \cmidrule(lr){2-3} \cmidrule(lr){4-5} 
         & \textit{Avg.} & \textit{Perf. Loss} & \textit{Avg.} & \textit{Comp. Rate\%} \\
        \midrule \midrule
        Original        & 54.2 & - & 8,603 & - \\
        \hdashline 
        Ours ($R_{\mathrm{c}}$)  & 53.5 & $\down{0.7}$ & 4,906 & $\up{43.0\%}$ \\
        Ours ($R_{\mathrm{marginal}}$)  & 51.2 & $\down{3.0}$ & 9,056 & $\down{5.3\%}$ \\
        \bottomrule
    \end{tabular*}
    \caption{Experiment results on Marginal Probability in \texttt{Qwen-2.5-7B-Instruct}.}
    \label{tab:marginal_prob.}
\end{table}
\subsubsection{Compression Rate Reward during RL}
We further examine the utility of incorporating a compression rate reward alongside the ConMax reward $R_{\mathrm{info}}(\hat{z})$. Explicit length penalties or rewards are widely used in various RL-based methods to incentivize brevity~\citep{arora2025training,team2025kimi}. We define this reward as the relative length reduction:
\begin{equation}
R_{\text{len}}(\hat{z}) = \frac{|z| - |\hat{z}|}{|z|}.
\end{equation}
This formulation directly quantifies the extent of compression. A positive reward is accrued when the compressed chain $\hat{z}$ is shorter than the original $z$, whereas a penalty (negative reward) is incurred if $\hat{z}$ exceeds the length of $z$. Theoretically, this encourages the model to eliminate redundant tokens while retaining essential information.

However, directly summing $R_{\text{c}}$ and $R_{\text{len}}$ is problematic due to scale discrepancies: $R_{\text{c}}$ typically consists of log-probabilities (negative values), while $R_{\text{len}}$ is a ratio ranging between 0 and 1 (or lower). To address this, we apply group-based normalization (whitening) to each reward component independently during the rollout phase. The normalized rewards, $R_{\text{c}}$ and $R_{\text{len}}$, are then combined using a weighting factor $\lambda$:
\begin{equation}
\tilde{R}(\hat{z}) = R_{\text{c}}(\hat{z}) + \lambda \cdot R_{\text{len}}(\hat{z}),
\end{equation}
where $\lambda$ modulates the trade-off between the informativeness of the trace and its conciseness. 

\begin{table}[!h]
    \small
    \renewcommand{\arraystretch}{1.22} 
    \centering
    % 调整了列间距，因为列数变少了，可以让表格更宽敞一些
    \setlength{\tabcolsep}{1pt} 
    
    % 定义列格式：左对齐的方法列，后面跟着两组居中的数据列
    \begin{tabular*}{\columnwidth}{@{\extracolsep{\fill}} l cc cc }
        \toprule
        \multirow{1}{*}{\textbf{Method}} & \multicolumn{2}{c}{\textbf{Accuracy~\%}~$\uparrow$} & \multicolumn{2}{c}{\textbf{Token Length}~$\downarrow$} \\
        \cmidrule(lr){2-3} \cmidrule(lr){4-5} 
         & \textit{Avg.} & \textit{Perf. Loss} & \textit{Avg.} & \textit{Comp. Rate\%} \\
        \midrule \midrule
        
        Original                & 54.2 & - & 8,603 & - \\
        \hdashline 
        Ours ($\lambda = 0$)    & 53.5 & $\down{1.3}$ & 4,906 & $\up{43.0\%}$ \\
        Ours ($\lambda = 0.01$) & 52.0 & $\down{2.2}$ & 4,880 & $\up{43.3\%}$ \\
        Ours ($\lambda = 0.05$) & 50.9 & $\down{3.3}$ & 5,185 & $\up{39.7\%}$ \\
        
        \bottomrule
    \end{tabular*}
    \caption{Experiment results on compression rate reward in \texttt{Qwen-2.5-7B-Instruct}.}
    \label{tab:leng_penalty.}
\end{table}

Table~\ref{tab:leng_penalty.} presents the experimental results of applying the compression rate reward to Qwen-2.5-7B. We tested small values of $\lambda=0.01$ and $\lambda=0.05$; however, $R_{\mathrm{len}}$ yielded no positive influence on performance and had a negligible effect on compression. We hypothesize that the explicit compression rate reward becomes functionally redundant, as the compression objective is already effectively internalized by the policy through the system prompt design (Figure~\ref{fig:system_prompt}). Furthermore, imposing an explicit length constraint may inadvertently restrict the policy's exploration, hindering the discovery of optimally compressed reasoning traces during the RL phase.

\section{Conclusion}
In this work, we introduce ConMax, a reinforcement learning framework that mitigates ``overthinking'' in Large Reasoning Models by compressing reasoning traces via reward-driven optimization. Our novel dual-confidence reward mechanism incorporates thinking confidence to ensure reasoning coherence and answer confidence to maintain predictive fidelity, thereby producing concise, high-quality training data. Empirical results on the Qwen-2.5 family, specifically the 7B model, demonstrate that ConMax reduces inference length by 43\% with a negligible 0.7\% performance drop. Overall, ConMax offers a robust solution for generating data-efficient corpora, paving the way for powerful yet computationally sustainable LRMs.

\section*{Limitations}
Despite the effectiveness of ConMax in enhancing verbose reasoning chains, several limitations remain to be explored in the future. First, our current evaluation is primarily conducted on mathematical and scientific reasoning tasks. While these results are encouraging, the generalizability of our approach to broader domains such as creative writing or code generation requires more extensive verification. Second, limited by computational constraints, our evaluation has not yet extended to ultra-large-scale models (e.g., 70B parameters or above) or architectures distinct from the Qwen family. Finally, ConMax's reward mechanism relies on a frozen auxiliary LRM to provide confidence estimation. Future work could explore self-supervised compression methods that do not rely on external auxiliary models.

\section*{Ethics Statement}
This study strictly adheres to community ethical guidelines. The datasets used (e.g., NuminaMath, AIME) are publicly available and focus on mathematical reasoning, ensuring they are free from personally identifiable information or discriminatory content. Our method, ConMax, is designed solely to improve the computational efficiency of Large Reasoning Models. We have carefully structured our training objectives to focus on logical compression, avoiding any generation of harmful or biased content. We foresee no direct negative societal impacts from this work. Furthermore, the backbone model and datasets used in this work are publicly available and legally permissible for research use.

% \section*{Ethics Statement}
% \zx{[complete this]}

%\clearpage

\bibliography{custom}
\appendix
\begin{figure*}[t]
    \centering
    \includegraphics[width=0.95\textwidth]{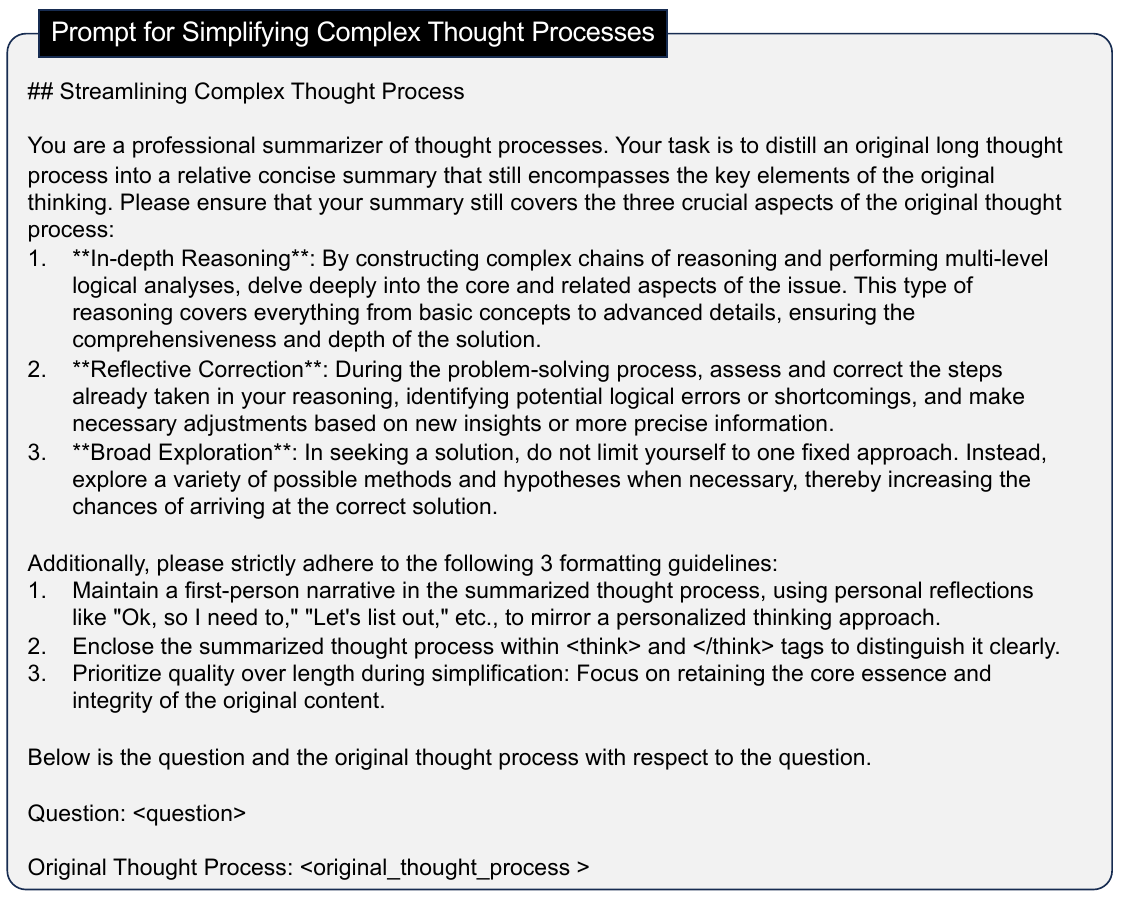}
    \caption{The system prompt to instruct the policy to compress verbose reasoning chains.}
    \label{fig:system_prompt}
\end{figure*}

\end{document}